\documentclass[nohyperref]{article}

\usepackage{microtype}
\usepackage{graphicx}
\usepackage{subfig}
\usepackage{booktabs} 

\usepackage{hyperref}

\usepackage[accepted]{pre-training}

\usepackage{amsmath}
\usepackage{amssymb}
\usepackage{mathtools}
\usepackage{amsthm}

\usepackage{verbatim}

\usepackage{algorithmic}
\usepackage{algorithm}

\usepackage{numprint}

\usepackage[capitalize,noabbrev]{cleveref}

\theoremstyle{plain}

\theoremstyle{definition}

\theoremstyle{remark}

\usepackage[textsize=small]{todonotes}  
\newcommand{\note}[1]{\todo[inline]{#1}}

\icmltitlerunning{On the Importance of Hyperparameters and Data Augmentation for SSL}

\begin{document}

\twocolumn[
\icmltitle{On the Importance of Hyperparameters and \\ Data Augmentation for Self-Supervised Learning}




\begin{icmlauthorlist}
\icmlauthor{Diane Wagner}{freiburg}
\icmlauthor{Fabio Ferreira}{freiburg}
\icmlauthor{Danny Stoll}{freiburg}
\icmlauthor{Robin Tibor Schirrmeister}{freiburg}
\icmlauthor{Samuel Müller}{freiburg}
\icmlauthor{Frank Hutter}{freiburg,bosch}
\end{icmlauthorlist}

\icmlaffiliation{freiburg}{Department of Computer Science, University of Freiburg, Freiburg, Germany}
\icmlaffiliation{bosch}{Bosch Center for Artificial Intelligence, Germany}

\icmlcorrespondingauthor{Diane Wagner}{wagnerd@cs.uni-freiburg.de}

\icmlkeywords{Self-Supervised Learning, Data Augmentation, Hyperparameters}

\vskip 0.3in
]



\printAffiliationsAndNotice{}  




\begin{abstract}
Self-Supervised Learning (SSL) has become a very active area of Deep Learning research where it is heavily used as a pre-training method for classification and other tasks. However, the rapid pace of advancements in this area comes at a price: training pipelines vary significantly across papers, which presents a potentially crucial confounding factor. Here, we show that, indeed, the choice of hyperparameters and data augmentation strategies can have a dramatic impact on performance. To shed light on these neglected factors and help maximize the power of SSL, we hyperparameterize these components and optimize them with Bayesian optimization, showing improvements across multiple datasets for the SimSiam SSL approach. Realizing the importance of data augmentations for SSL, we also introduce a new automated data augmentation algorithm, \textit{GroupAugment}, which considers groups of augmentations and optimizes the sampling across groups. In contrast to algorithms designed for supervised learning, GroupAugment achieved consistently high linear evaluation accuracy across all datasets we considered. Overall, our results indicate the importance and likely underestimated role of data augmentation for SSL.
\end{abstract}

\section{Introduction}

Self-supervised learning (SSL) has seen an explosion of research interest in recent years, with significant progress using SSL as a pre-training method for classification \citep{grill2020bootstrap, chen2021exploring, chen2020simple, he2020momentum, mocov2}.
A large variety of SSL methods have been developed, for example using different optimization paradigms \citep{cpc-vandenoord,pathak2016context}, different objective functions \citep{gidaris-iclr18a, doersch-iccv15a, zhang-eccv16a}, or different data modalities \citep{radford2021learning}.


An aspect of SSL performance that is less researched is the effect of other choices, such as the training hyperparameters or the augmentation strategy. To shed light on these neglected factors, we use Bayesian optimization \citep{mockus-tgo78a,shahriari-ieee16a} to search for configurations of the SimSiam SSL algorithm \citep{chen2021exploring} on CIFAR-10, CIFAR-100 \citep{krizhevsky-tech09a}, and the medical dataset DermaMNIST \citep{medmnistv1,medmnistv2}. We consider, on the one hand, a search over training hyperparameters and, on the other hand, a search over data augmentation strategies. Among other findings, our results suggest the importance of data augmentation for SSL.


Motivated by the 
apparent importance of data augmentation for SSL, 
we then develop a new automated data augmentation algorithm, \emph{GroupAugment}, that covers a more diverse space of augmentation strategies than existing methods and can, 
e.g., design augmentation strategies that resemble manually-designed SSL augmentation strategies. 

In summary, our main contributions are:
\begin{itemize}
    \item We study the effect of training hyperparameters and augmentation strategies for the SimSiam SSL approach (Section~\ref{sec:study}). Our results indicate the importance of data augmentation for SSL.
    \item We introduce the automated data augmentation algorithm GroupAugment and demonstrate its high performance for SSL (Section~\ref{sec:groupaugment}).
\end{itemize}



\section{Background and Related Work}\label{sec:related}

\paragraph{Self-Supervised Learning}  
The most cited works in Self-Supervised Learning, such as SimCLR, BYOL, SimSiam, MoCo, or DINO \citep{grill2020bootstrap, chen2021exploring, chen2020simple, he2020momentum, mocov2} all apply a similar or identical data augmentation protocol (random horizontal flip, color distortion, and nowadays also Gaussian blur and solarization). Most relevant to our work are the findings of \citet{grill2020bootstrap} and \citet{chen2020simple}, who identify and address the sensitivity to choosing color distortions in their methods. Moreover, \citet{chen2020simple} identified that a sophisticated supervised data augmentation strategy does not perform better than simple cropping with strong color distortion in the self-supervised setting. These observations motivate the contributions regarding data augmentation for SSL, which we will revisit next.  






\paragraph{Data Augmentation Algorithms} 
In supervised learning, data augmentation algorithms usually outperform manually selected augmentation strategies: Three algorithms that rely on randomly sampling augmentations from a fixed set of augmentations are TrivialAugment \citep{muller-iccv21} and SmartSamplingAugment \citep{a15050165}. Other algorithms, such as AutoAugment \citep{cubuk-cvpr19}, RandAugment \citep{cubuk-cvpr20}, and SmartAugment \citep{a15050165} perform a gradient-free search in the space of augmentation policies.
%
It is also possible to apply gradient-based optimization to meta-learn pre-training hyperparameters \citep{raghu2021meta} for ECG data. In SelfAugment \citep{reed2021selfaugment}, the authors bootstrap from the correlation between supervised and self-supervised evaluation performance to incorporate a rotation task for generating augmentation policies efficiently. 
SelfAugment is qualitatively different from the previous approaches in that it does not optimize for general downstream performance but rotation task performance. Additionally, it will likely not include rotations into the selected augmentation policy.


In contrast to these, our approach \emph{GroupAugment} (Section~\ref{sec:groupaugment}) covers a more diverse space of augmentation strategies than existing methods and can, e.g., design augmentation strategies that resemble manually-designed SSL augmentation strategies. It generalizes existing approaches by optimizing group-specific sampling probabilities, the number of group-specific augmentations, and the total number of augmentations applied while imposing no limitations on the set of augmentations such as SelfAugment.

\section{Study on the Importance of Hyperparameters and Data Augmentation}\label{sec:study}

We study the following research questions:
\vspace{-0.1cm}
\begin{itemize}
\item What role does data augmentation play in SSL, and can better data augmentation strategies lead to better performance?
\item Which hyperparameters may be notorious for resulting in model collapse when set incorrectly?
\item Which hyperparameters are important to optimize in SSL to achieve good performance and outperform baselines?
\end{itemize}


\subsection{Study Design} 

To answer the presented questions, we conducted the study described below.

\paragraph{Models, Datasets and Hyperparameter Search Spaces}
We perform all our experiments on the CIFAR-10, CIFAR-100 and DermaMNIST datasets \citep{krizhevsky-tech09a, medmnistv1,medmnistv2} and use the SimSiam \citep{chen2021exploring} approach with the ResNet-18 \citep{he-ieee16} architecture. We optimize a wide range of training pipeline hyperparameters (which we refer to as \emph{Training Hypers}) such as the learning rate, warmup and weight decay, and optimizer, as well as data augmentation hyperparameters (which we refer to as \emph{Augmentations}) involving magnitudes and probabilities of image distortions. Please see Appendix~\ref{sec:appendix-search-spaces} for more details on our search spaces. We point out that all hyperparameter search spaces of the baselines are chosen identically across all datasets except for the pre-training epochs, where we used 800 for CIFAR-10/100 and 100 for DermaMNIST. Lastly, we report how the train, validation, and test splits were chosen in Appendix \ref{apx:priors}.

\paragraph{Search Algorithm} To optimize over the search spaces listed above, we use Bayesian optimization (BO) \citep{mockus-tgo78a} with expert priors \citep{hvarfner2021pibo} as implemented by \citet{stoll2022neps}. For details on the chosen priors see Appendix~\ref{apx:priors}.


\paragraph{Performance Evaluation}
All reported performance values are based on the standard linear evaluation protocol \citep{dalal, grill2020bootstrap} from the SSL literature that trains a linear classifier on top of the frozen ResNet backbone weights.


\subsection{Results}
\begin{table*}[t!]
    \centering
        \caption{Mean test accuracy $[\%]$ for SimSiam and its tuned variants in the linear evaluation protocol. We report the mean and standard error across five seeds. ($\dagger$) The original SimSiam result of $91.8\%$ was achieved using early stopping on the test set.}
    \begin{tabular}{llll}
    \toprule
    Approach & DermaMNIST & CIFAR-10 &   CIFAR-100 \\
    \midrule
    SimSiam \citep{chen2021exploring}               &   $66.2$ $\pm 0.3$ & $91.6$  $\pm 0.1^\dagger$ & $65.6 $ $\pm 0.2$ \\
    SimSiam Tuned Training Hypers          & $66.5$ $\pm 0.1$ & $91.6$ $\pm 0.1$ &  $64.9$ $\pm 0.2$ \\
    SimSiam Tuned Augmentations  & $\textbf{67.2}$ $ \pm 0.4$ & $\textbf{92.7}$ $\pm 0.1$ & $\textbf{67.9}$ $ \pm 0.3$\\
    \bottomrule
    \end{tabular}
    \label{table:simsiam-results}
\end{table*}


\paragraph{Hyperparameters vs Data Augmentation Strategy} Table \ref{table:simsiam-results} shows that (a) optimizing six training hyperparameters (detailed in Table \ref{table:search-space-traininghypers} in the appendix) of SimSiam only lead to marginal improvements or even deterioration of performance (due to differences in validation and test split); (b) optimizing the data augmentation strategy lead to consistent significant performance improvements (at least 1\%, and up to 2.3\% for CIFAR-100).
This shows that SimSiam's training hyperparameters were already very well-tuned.

\begin{figure}[ht!]
    \centering
    \includegraphics[width=0.35\textwidth]{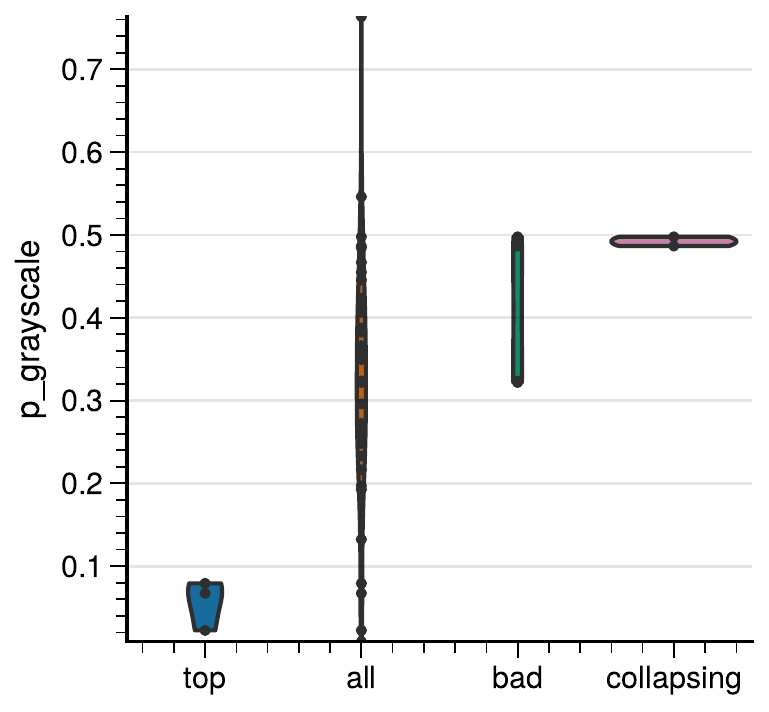}
    \caption{Density estimates for the probability of applying a grayscale augmentation as sampled in our optimization for SimSiam's augmentation strategy on CIFAR-100. We show density estimates for the best and worst-performing configurations (top and bad), all configurations (all), and configurations leading to collapsing.}
    \label{fig:p_grayscale}
\end{figure}

%
\paragraph{How to Avoid Collapsing} \citet{chen2021exploring} already analyzed which factors, e.g., stop-gradient, can cause collapsing solutions for SimSiam. A collapsing solution is an undesired solution where all the outputs collapse to a constant vector. We continue this study and give insights into which hyperparameters may cause collapsing if chosen suboptimally. While we observed collapsing solutions in some of our experiments on the CIFAR-10 and CIFAR-100 datasets, for DermaMNIST, we surprisingly observed no collapsing solutions. We show violin plots in Appendix~\ref{sec:hyperparameter-analysis} giving some insights on which hyperparameters may be responsible for collapsing solutions. For example, Figure~\ref{fig:p_grayscale}, an excerpt from Appendix \ref{sec:hyperparameter-analysis}, indicates that a probability of applying grayscale to CIFAR-100 around $0.5$ might cause collapsing solutions.

\paragraph{Importance of Hyperparameters} In the following, we study the importance of individual hyperparameters for achieving good performance. First, we compare the manually designed hyperparameter settings from \citet{chen2021exploring} to the optimized settings found in our study. As \citet{chen2021exploring} do not report results on CIFAR-100 and DermaMNIST, we focus on CIFAR-10 here. For training hyperparameters, we do not observe significant differences in performance (cf. Table~\ref{table:simsiam-results}), and the settings found by \cite{chen2021exploring} seem to be quite optimal. For the parameterized data augmentation config space, we observe that the grayscale probability hyperparameter should be set half as high as in the baseline and that the saturation strength from the color jitter needs to be higher than in the baseline. Further, adding solarize seems important, which is also observed by \citet{grill2020bootstrap}. The hyperparameter importance study in Table~\ref{table:fanova-cifar10-dataaugmentation} also supports the importance of the above parameters.
Additionally, Figure~\ref{fig:cifar10-individual-simsiamdataaug} suggests that for the above parameters, good and bad configurations differ, supporting our findings. 
Moreover, our results also give insights into how individual hyperparameters should be optimally set. As an example, Figure \ref{fig:p_grayscale} shows how the probability of applying grayscale should be optimally set for the CIFAR-100 dataset. In the best-performing configurations, this probability hyperparameter has been sampled below $0.1$ and in bad-performing configurations above $0.3$. We show violin plots of all the other individual hyperparameters of the different config spaces and datasets in Appendix \ref{sec:hyperparameter-analysis}.

%
%

\begin{table}[ht]
    \centering
    \caption{Hyperparameter importances for the SimSiam data augmentation space for CIFAR-10. We utilized fANOVA \citep{hutter2014efficient}, which quantifies the contribution of individual hyperparameters to the overall variance in performance. We distinguish between the general importance and the importance of best configurations. Higher importance values denote a higher performance responsibility.}
    \begin{tabular}{lcc}
    \toprule
    Hyperparameter & Across all & Across best \\
    \midrule
    brightness strength &  1 &  2 \\
    contrast strength &  2 &  2 \\
    hue strength &  8 &  3 \\
    p colorjitter &  2 &  2 \\
    p grayscale & 16 & 32 \\
    p horizontal flip &  8 &  2 \\
    p solarize & 30 & 26 \\
    saturation strength & 23 &  7 \\
    solarize threshold &  5 &  2 \\
    \bottomrule
    \end{tabular}
    \label{table:fanova-cifar10-dataaugmentation}
\end{table}

\vspace{-0.5cm}

\section{GroupAugment}\label{sec:groupaugment}


In this section, we introduce \textit{GroupAugment}, an automated data augmentation algorithm that operates on groups of augmentations (such as color or quality transformations) and designs sampling strategies over these groups. Further, we present an empirical study where we find that, in contrast to the data augmentation algorithms designed for supervised learning, GroupAugment robustly outperforms the baseline in all the settings we analyzed.

\subsection{Algorithm}

\paragraph{Augmentation Sampling} Given some groups of data augmentations $\{ g_i \}$, GroupAugment uses one global hyperparameter and two sets of group-specific hyperparameters to create a list of augmentation sequences that will be consecutively applied to an image. The global hyperparameter $T$ determines the number of augmentation sequences in that list. The group-specific hyperparameter $P_{g_i}$ determines the probability that augmentation group~$g_i$ is chosen to create the following augmentation sequence. $N_{g_i}$ determines the number of augmentations to sample uniformly without replacement to form an augmentation sequence for augmentation group~$g_i$. We provide pseudocode for a GroupAugment policy in Algorithm~\ref{alg:groupaugment_algorithm}.

\begin{algorithm}[ht!]
   \caption{A GroupAugment policy applied to an image.}
   \label{alg:groupaugment_algorithm}
    \begin{algorithmic}
       \STATE {\bfseries Input:}
       Image $I$, augmentation groups $\{ g_i \}$, \\
       \hspace{.96cm} group-specific sampling probabilities $\{ P_{g_i} \}$, \\ 
       \hspace{.96cm} group-specific \#augmentations $\{ N_{g_i} \}$, \\ 
       \hspace{.96cm} total \#group-samples $T$\\
       \STATE{Initialize empty list of augmentation sequences $A$}
       \FOR{augmentation sequence $1, \ldots, T$}
       \STATE{Sample group $g$ according to $\{ P_{g_i} \}$} \\
       \STATE{Sample $N_{g}$ augmentations from group $g$} \\
       \STATE{Append augmentations to $A$} \\
       \ENDFOR \\
       \STATE{Apply sampled augmentation sequences $A$ to $I$}
\end{algorithmic}
\end{algorithm}

\paragraph{Novelty of Group Sampling} While \citet{a15050165} explored optimizing the parameters of color and geometric augmentation groups, GroupAugment generalizes this notion to \emph{any} set of augmentation groups and searches over more general spaces of sampling strategies. In our study, we instantiate GroupAugment with five groups: color, geometric, non-rigid, quality, and exotic. See Table~\ref{table:groupaugment_augmentations} in Appendix~\ref{sec:appendix-groupaugment} for the specific augmentations we used and Table~\ref{table:groupaugment-search-space} in Appendix~\ref{sec:appendix-search-spaces} for a detailed description of GroupAugment's search space.


\paragraph{Search Algorithm} To optimize over the resulting search space, we use the same Bayesian optimization (BO) approach with expert priors as above. Further, we normalize the sampled probabilities, as BO samples each group probability individually, and therefore, they do not necessarily add up to~1.

\subsection{Comparisons and Results}
We compare GroupAugment to several automated data augmentation algorithms. On the one hand, we observe that applying standard automated data augmentation algorithms, e.g., RandAugment, to the self-supervised learning setting worsens the results for most of our analyzed datasets, although those algorithms perform well in the supervised-learning setting (see Table~\ref{table:simsiam-results-augmentations}). This observation is in line with the finding of \citet{chen2020simple} that a sophisticated supervised data augmentation strategy did not perform better than simple cropping with strong color distortion in the SSL setting. On the other hand, contrary to standard supervised-learning data augmentation algorithms, GroupAugment robustly outperforms the baseline in all the settings we analyzed. We give more details on the experimental settings in Appendix~\ref{apx:priors}.\\
Further, we observe a performance improvement over the baseline when tuning its data augmentation strategy's magnitudes and application probabilities with the same computational budget we used for the GroupAugment search space. However, as GroupAugment outperforms the tuned SimSiam data augmentation for most of our results, allowing stronger parameterization, we recommend optimizing the data augmentation with GroupAugment, especially for datasets having no intuition about which data augmentation might be helpful.

\begin{table}[t!]
\scriptsize
\setlength{\tabcolsep}{0.03cm}

    \centering
        \caption{Mean test accuracy for different algorithms in the linear evaluation protocol. We report the mean and standard error across five seeds for methods we ran. For each dataset, we bold the two best accuracies and underline scores that outperform the SimSiam baseline. ($\dagger$) The original SimSiam result of $91.8\%$ was achieved using early stopping on the test set. ($\ddagger$) SelfAugment and SelfRandAugment were evaluated using Resnet50 and not Resnet18 as other methods. \citet{reed2021selfaugment} report results for multiple instantiations of SelfAugment.}
    \begin{tabular}{lccc}
    \toprule
    Approach & DermaMNIST & CIFAR-10 &   CIFAR-100 \\
    \midrule
    SimSiam \citep{chen2021exploring}               &   $66.2$ $\pm 0.3$ & $91.6$  $\pm 0.1^\dagger$ & $65.6 $ $\pm 0.2$ \\
    \midrule
    SelfRandAugment \citep{reed2021selfaugment}      &  -      & $90.3^\ddagger$ & - \\
    SelfAugment \citep{reed2021selfaugment}         &  -  & $87.5-92.6^\ddagger$  & - \\
    RandAugment \citep{cubuk-cvpr20}        & \underline{$\textbf{68.8}$} $\pm 0.3$ & $89.9 \pm 0.0$  & $59.3 \pm 0.5$\\
    SmartAugment \citep{a15050165}           & \underline{$67.5$} $\pm 0.1$ & $89.8 \pm 0.1$  & $59.8 \pm 0.1$ \\
    TrivialAugment \citep{muller-iccv21} & \underline{$67.7$} $\pm 0.5$  & $89.4 \pm 0.1$ & $59.1\pm 0.2$ \\
    \midrule
    Tuned SimSiam Augmentations (our)  & \underline{$67.2$} $ \pm 0.4$ &     \underline{$\textbf{92.7}$} $ \pm 0.1$ & \underline{$\textbf{67.9}$} $ \pm 0.3$\\
    GroupAugment (our)  & \underline{$\textbf{68.0}$} $ \pm 0.3$   & \underline{$\textbf{93.0}$} $ \pm 0.1$ & \underline{$\textbf{66.3}$} $\pm 0.4$ \\
    \bottomrule
    \end{tabular}
    \label{table:simsiam-results-augmentations}
\end{table}


\section{Conclusion and Limitations}



While SimCLR and BYOL have analyzed the role of data augmentation, our results show that it is beneficial to analyze it in much greater detail. In summary, we provide evidence for the underestimated role of data augmentation for SSL and present a novel automated data augmentation algorithm, GroupAugment, which outperforms vanilla-SimSiam across all datasets we study.

\paragraph{Limitations}
We conducted our study on CIFAR-10, CIFAR-100 \citep{krizhevsky-tech09a}, and DermaMNIST \citep{medmnistv1,medmnistv2}. More datasets should be considered to gather more insights and strengthen our experimental conclusions. In particular, results on ImageNet \citep{deng-cvpr09} and more medical datasets are of interest. Further, as automated data augmentation (e.g., our GroupAugment) optimizes the augmentation strategy on a validation set, out-of-distribution test sets can pose a challenge if the validation set is in-distribution.

\section*{Acknowledgements}
We acknowledge funding by Robert Bosch GmbH, by the Deutsche Forschungsgemeinschaft (DFG, German Research Foundation) under grant number 417962828, by European Research Council (ERC) Consolidator Grant ``Deep Learning 2.0'' (grant no.\ 101045765), and by BrainLinks-BrainTools which was funded by the German Research Foundation (DFG, grant no. EXC 1086) and is currently funded by the Federal Ministry of Economics, Science and Arts of Baden W\"urttemberg within the sustainability program for projects of the excellence initiative. Funded by the European Union. Views and opinions expressed are however those of the author(s) only and do not necessarily reflect those of the European Union or the ERC. Neither the European Union nor the ERC can be held responsible for them.
\begin{center}\includegraphics[width=0.3\textwidth]{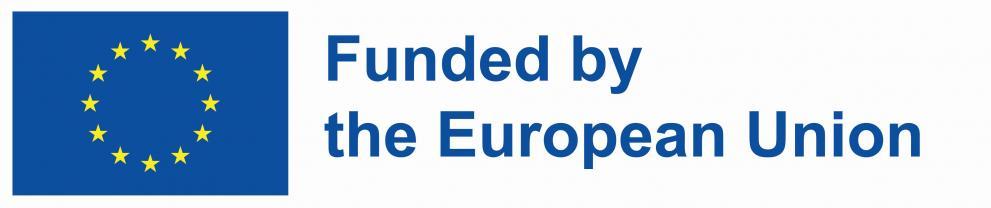}\end{center}

\newpage
\bibliography{bibtex/lib.bib,bibtex/local.bib,bibtex/proc.bib,bibtex/strings.bib}

\begin{thebibliography}{29}
\providecommand{\natexlab}[1]{#1}
\providecommand{\url}[1]{\texttt{#1}}
\expandafter\ifx\csname urlstyle\endcsname\relax
  \providecommand{\doi}[1]{doi: #1}\else
  \providecommand{\doi}{doi: \begingroup \urlstyle{rm}\Url}\fi

\bibitem[Buslaev et~al.(2020)Buslaev, Iglovikov, Khvedchenya, Parinov,
  Druzhinin, and Kalinin]{buslaev2020albumentations}
Buslaev, A., Iglovikov, V.~I., Khvedchenya, E., Parinov, A., Druzhinin, M., and
  Kalinin, A.~A.
\newblock Albumentations: Fast and flexible image augmentations.
\newblock \emph{Information}, 11\penalty0 (2), 2020.
\newblock ISSN 2078-2489.
\newblock \doi{10.3390/info11020125}.
\newblock URL \url{https://www.mdpi.com/2078-2489/11/2/125}.

\bibitem[Chen et~al.(2020{\natexlab{a}})Chen, Kornblith, Norouzi, and
  Hinton]{chen2020simple}
Chen, T., Kornblith, S., Norouzi, M., and Hinton, G.
\newblock A simple framework for contrastive learning of visual
  representations.
\newblock In \emph{International conference on machine learning}, pp.\
  1597--1607. PMLR, 2020{\natexlab{a}}.

\bibitem[Chen \& He(2021)Chen and He]{chen2021exploring}
Chen, X. and He, K.
\newblock Exploring simple siamese representation learning.
\newblock In \emph{Proceedings of the IEEE/CVF Conference on Computer Vision
  and Pattern Recognition}, pp.\  15750--15758, 2021.

\bibitem[Chen et~al.(2020{\natexlab{b}})Chen, Fan, Girshick, and He]{mocov2}
Chen, X., Fan, H., Girshick, R., and He, K.
\newblock Improved baselines with momentum contrastive learning.
\newblock \emph{arXiv preprint arXiv:2003.04297}, 2020{\natexlab{b}}.

\bibitem[Cubuk et~al.(2019)Cubuk, Zoph, Mane, Vasudevan, and Le]{cubuk-cvpr19}
Cubuk, E., Zoph, B., Mane, D., Vasudevan, V., and Le, Q.
\newblock {Autoaugment}: Learning augmentation strategies from data.
\newblock In \emph{Proceedings of the {IEEE/CVF} Conference on Computer Vision
  and Pattern Recognition}, pp.\  113--123, 2019.

\bibitem[Cubuk et~al.(2020)Cubuk, Zoph, Shlens, and Le]{cubuk-cvpr20}
Cubuk, E., Zoph, B., Shlens, J., and Le, Q.
\newblock Randaugment: Practical automated data augmentation with a reduced
  search space.
\newblock In \emph{Proceedings of the {IEEE/CVF} Conference on Computer Vision
  and Pattern Recognition Workshops}, pp.\  702--703, 2020.

\bibitem[Dalal \& Triggs(2005)Dalal and Triggs]{dalal}
Dalal, N. and Triggs, B.
\newblock Histograms of oriented gradients for human detection.
\newblock In \emph{CVPR'05}, volume~1, pp.\  886--893 vol. 1, 2005.

\bibitem[Deng et~al.(2009)Deng, Dong, Socher, Li, Li, and Fei-Fei]{deng-cvpr09}
Deng, J., Dong, W., Socher, R., Li, L., Li, K., and Fei-Fei, L.
\newblock {ImageNet: A Large-Scale Hierarchical Image Database}.
\newblock In \emph{Proceedings of the International Conference on Computer
  Vision and Pattern Recognition ({CVPR}'09)}, pp.\  248--255, 2009.

\bibitem[Doersch et~al.(2015)Doersch, Gupta, and Efros]{doersch-iccv15a}
Doersch, C., Gupta, A., and Efros, A.~A.
\newblock Unsupervised visual representation learning by context prediction.
\newblock In \emph{Proceedings of the 18th International Conference on Computer
  Vision ({ICCV}'15)}, pp.\  1422--1430, 2015.

\bibitem[Giradis et~al.(2018)Giradis, Singh, and Komodakis]{gidaris-iclr18a}
Giradis, S., Singh, P., and Komodakis, N.
\newblock Unsupervised representation learning by predicting image rotations.
\newblock In \emph{Proceedings of the International Conference on Learning
  Representations ({ICLR}'18)}, 2018.

\bibitem[Grill et~al.(2020)Grill, Strub, Altch{\'e}, Tallec, Richemond,
  Buchatskaya, Doersch, Avila~Pires, Guo, Gheshlaghi~Azar,
  et~al.]{grill2020bootstrap}
Grill, J.-B., Strub, F., Altch{\'e}, F., Tallec, C., Richemond, P.,
  Buchatskaya, E., Doersch, C., Avila~Pires, B., Guo, Z., Gheshlaghi~Azar, M.,
  et~al.
\newblock Bootstrap your own latent-a new approach to self-supervised learning.
\newblock \emph{Advances in Neural Information Processing Systems},
  33:\penalty0 21271--21284, 2020.

\bibitem[He et~al.(2016)He, Zhang, Ren, and Sun]{he-ieee16}
He, K., Zhang, X., Ren, S., and Sun, J.
\newblock Deep residual learning for image recognition.
\newblock In \emph{Proceedings of the {IEEE} conference on computer vision and
  pattern recognition}, pp.\  770--778, 2016.

\bibitem[He et~al.(2020)He, Fan, Wu, Xie, and Girshick]{he2020momentum}
He, K., Fan, H., Wu, Y., Xie, S., and Girshick, R.
\newblock Momentum contrast for unsupervised visual representation learning.
\newblock In \emph{Proceedings of the IEEE/CVF conference on computer vision
  and pattern recognition}, pp.\  9729--9738, 2020.

\bibitem[Hutter et~al.(2014)Hutter, Hoos, and
  Leyton-Brown]{hutter2014efficient}
Hutter, F., Hoos, H., and Leyton-Brown, K.
\newblock An efficient approach for assessing hyperparameter importance.
\newblock In \emph{International conference on machine learning}, pp.\
  754--762. PMLR, 2014.

\bibitem[Hvarfner et~al.(2021)Hvarfner, Stoll, Souza, Lindauer, Hutter, and
  Nardi]{hvarfner2021pibo}
Hvarfner, C., Stoll, D., Souza, A., Lindauer, M., Hutter, F., and Nardi, L.
\newblock $\pi$bo: Augmenting acquisition functions with user beliefs for
  bayesian optimization.
\newblock In \emph{Proceedings of the International Conference on Learning
  Representations ({ICLR}'21)}, 2021.
\newblock Published online: \url{iclr.cc}.

\bibitem[Krizhevsky(2009)]{krizhevsky-tech09a}
Krizhevsky, A.
\newblock Learning multiple layers of features from tiny images.
\newblock Technical report, University of Toronto, 2009.

\bibitem[Mockus et~al.(1978)Mockus, Tiesis, and Zilinskas]{mockus-tgo78a}
Mockus, J., Tiesis, V., and Zilinskas, A.
\newblock The application of {B}ayesian methods for seeking the extremum.
\newblock \emph{Towards Global Optimization}, 2\penalty0 (117-129), 1978.

\bibitem[M\"uller \& Hutter(2021)M\"uller and Hutter]{muller-iccv21}
M\"uller, S.~G. and Hutter, F.
\newblock Trivialaugment: Tuning-free yet state-of-the-art data augmentation.
\newblock In \emph{Proceedings of the IEEE/CVF International Conference on
  Computer Vision (ICCV)}, pp.\  774--782, October 2021.

\bibitem[Negassi et~al.(2022)Negassi, Wagner, and Reiterer]{a15050165}
Negassi, M., Wagner, D., and Reiterer, A.
\newblock Smart(sampling)augment: Optimal and efficient data augmentation for
  semantic segmentation.
\newblock \emph{Algorithms}, 15\penalty0 (5), 2022.
\newblock ISSN 1999-4893.
\newblock \doi{10.3390/a15050165}.
\newblock URL \url{https://www.mdpi.com/1999-4893/15/5/165}.

\bibitem[Pathak et~al.(2016)Pathak, Krahenbuhl, Donahue, Darrell, and
  Efros]{pathak2016context}
Pathak, D., Krahenbuhl, P., Donahue, J., Darrell, T., and Efros, A.~A.
\newblock Context encoders: Feature learning by inpainting.
\newblock In \emph{Proceedings of the IEEE conference on computer vision and
  pattern recognition}, pp.\  2536--2544, 2016.

\bibitem[Radford et~al.(2021)Radford, Kim, Hallacy, Ramesh, Goh, Agarwal,
  Sastry, Askell, Mishkin, Clark, et~al.]{radford2021learning}
Radford, A., Kim, J.~W., Hallacy, C., Ramesh, A., Goh, G., Agarwal, S., Sastry,
  G., Askell, A., Mishkin, P., Clark, J., et~al.
\newblock Learning transferable visual models from natural language
  supervision.
\newblock In \emph{International Conference on Machine Learning}, pp.\
  8748--8763. PMLR, 2021.

\bibitem[Raghu et~al.(2021)Raghu, Lorraine, Kornblith, McDermott, and
  Duvenaud]{raghu2021meta}
Raghu, A., Lorraine, J., Kornblith, S., McDermott, M., and Duvenaud, D.~K.
\newblock Meta-learning to improve pre-training.
\newblock \emph{Advances in Neural Information Processing Systems}, 34, 2021.

\bibitem[Reed et~al.(2021)Reed, Metzger, Srinivas, Darrell, and
  Keutzer]{reed2021selfaugment}
Reed, C.~J., Metzger, S., Srinivas, A., Darrell, T., and Keutzer, K.
\newblock Selfaugment: Automatic augmentation policies for self-supervised
  learning.
\newblock In \emph{Proceedings of the IEEE/CVF Conference on Computer Vision
  and Pattern Recognition}, pp.\  2674--2683, 2021.

\bibitem[Shahriari et~al.(2016)Shahriari, Swersky, Wang, Adams, and
  de~Freitas]{shahriari-ieee16a}
Shahriari, B., Swersky, K., Wang, Z., Adams, R., and de~Freitas, N.
\newblock Taking the human out of the loop: {A} review of {B}ayesian
  optimization.
\newblock \emph{Proceedings of the {IEEE}}, 104\penalty0 (1):\penalty0
  148--175, 2016.

\bibitem[Stoll et~al.(2022)Stoll, Schrodi, Janowski, Mallik, Théophane, and
  Hutter]{stoll2022neps}
Stoll, D., Schrodi, S., Janowski, M., Mallik, N., Théophane, V., and Hutter,
  F.
\newblock {Neural Pipeline Search (NEPS)}, May 2022.
\newblock URL \url{https://github.com/automl/neps}.

\bibitem[van~den Oord et~al.(2018)van~den Oord, Li, and
  Vinyals]{cpc-vandenoord}
van~den Oord, A., Li, Y., and Vinyals, O.
\newblock Representation learning with contrastive predictive coding.
\newblock \emph{CoRR}, abs/1807.03748, 2018.

\bibitem[Yang et~al.(2021{\natexlab{a}})Yang, Shi, and Ni]{medmnistv1}
Yang, J., Shi, R., and Ni, B.
\newblock Medmnist classification decathlon: A lightweight automl benchmark for
  medical image analysis.
\newblock In \emph{IEEE 18th International Symposium on Biomedical Imaging
  (ISBI)}, pp.\  191--195, 2021{\natexlab{a}}.

\bibitem[Yang et~al.(2021{\natexlab{b}})Yang, Shi, Wei, Liu, Zhao, Ke, Pfister,
  and Ni]{medmnistv2}
Yang, J., Shi, R., Wei, D., Liu, Z., Zhao, L., Ke, B., Pfister, H., and Ni, B.
\newblock Medmnist v2: A large-scale lightweight benchmark for 2d and 3d
  biomedical image classification.
\newblock \emph{arXiv preprint arXiv:2110.14795}, 2021{\natexlab{b}}.

\bibitem[Zhang et~al.(2016)Zhang, Isola, and Efros]{zhang-eccv16a}
Zhang, R., Isola, P., and Efros, A.~A.
\newblock Colorful image colorization.
\newblock In \emph{12th European Conference on Computer Vision ({ECCV}'16)},
  pp.\  649--666, 2016.

\end{thebibliography}
\bibliographystyle{icml2022}

\newpage
\appendix
\onecolumn

\section{Search Spaces} \label{sec:appendix-search-spaces}

\begin{table}[ht!]
    \centering
    \caption{SimSiam data augmentation search space.}
    \begin{tabular}{lccccc}
    \toprule
    Hyperparameter & Type & Range & Log-Prior & Default \\
    \midrule
    p\_colorjitter & Float & [0, 1] & False & 0.8 \\
    p\_grayscale & Float & [0, 1] & False & 0.2 \\
    p\_horizontal\_flip & Float & [0, 1] & False & 0.5 \\
    p\_solarize & Float & [0, 1] & False & 0.2 \\
    brightness\_strength & Float & [0, 1.5] & False & 0.4 \\
    contrast\_strength & Float & [0, 1.5] & False & 0.4 \\
    saturation\_strength & Float & [0, 1.5] & False & 0.4 \\
    hue\_strength & Float & [0, 0.5] & False & 0.1 \\
    solarize\_threshold & Integer & [0, 255] & False & 127 \\
    \bottomrule
    \end{tabular}
    \label{table:search-space-simisiamaugs}
\end{table}

\begin{table}[ht!]
    \centering
    \caption{SimSiam training hyperparameters search space.}
    \begin{tabular}{lccccc}
    \toprule
    Hyperparameter & Type & Range & Log-Prior & Default \\
    \midrule
    learning\_rate & Float & [0.003, 0.3] & True & 0.03 \\
    warmup\_epochs & Integer & [0, 80] & False & 0 \\
    warmup\_multiplier & Float & [1.0, 3.0] & False & 1.0 \\
    optimizer & Categorical & \{AdamW, SGD, LARS\} & False & SGD \\
    weight\_decay\_start & Float & [\numprint{5e-6}, \numprint{5e-2}] & True & \numprint{5e-4} \\
    weight\_decay\_end & Float & [\numprint{5e-6}, \numprint{5e-2}] & True & \numprint{5e-4} \\
    \bottomrule
    \end{tabular}
    \label{table:search-space-traininghypers}
\end{table}

\begin{table}[ht!]
    \centering
    \caption{GroupAugment search space.}
    \begin{tabular}{lccccc}
    \toprule
    Hyperparameter & Type & Range & Log-Prior & Default \\
    \midrule
    p\_color\_transformations & Float & [0, 1] & False & 0.5 \\
    p\_geometric\_transformations & Float & [0, 1] & False & 0.5 \\
    p\_non\_rigid\_transformations & Float & [0, 1] & False & 0.0 \\
    p\_quality\_transformations & Float & [0, 1] & False & 0.0 \\ p\_exotic\_transformations & Float & [0, 1] & False & 0.0 \\
    num\_color\_transformations & Integer & [1, 5] & False & 1 \\
    num\_geometric\_transformations & Integer & [1, 2] & False & 1 \\
    num\_non\_rigid\_transformations & Integer & [1, 3] & False & 1 \\
    num\_quality\_transformations & Integer & [1, 2] & False & 1 \\ num\_exotic\_transformations & Integer & [1, 2] & False & 1 \\
    num\_total\_group\_samples & Integer & [1, 5] & False & 1 \\
    \bottomrule
    \end{tabular}
    \label{table:groupaugment-search-space}
\end{table}

\begin{table}[ht!]
    \centering
    \caption{RandAugment search space.}
    \begin{tabular}{lccccc}
    \toprule
    Hyperparameter & Type & Range & Log-Prior & Default \\
    \midrule
    num\_ops & Integer & [1, 15] & False & 3 \\
    magnitude & Integer & [0, 30] & False & 4 \\
    \bottomrule
    \end{tabular}
    \label{table:randaugment-config-spaces}
\end{table}

\begin{table}[ht!]
    \centering
    \caption{SmartAugment search space.}
    \begin{tabular}{lccccc}
    \toprule
    Hyperparameter & Type & Range & Log-Prior & Default \\
    \midrule
    num\_col\_ops & Integer & [1, 9] & False & 2 \\
    num\_geo\_ops & Integer & [1, 5] & False & 1 \\
    col\_magnitude & Integer & [0, 30] & False & 4 \\
    geo\_magnitude & Integer & [0, 30] & False & 4 \\
    p\_apply\_ops & Float & [0, 1] & False & 1 \\
    \bottomrule
    \end{tabular}
    \label{table:smartaugment-config-spaces}
\end{table}

\newpage
\section{Experimental Details}\label{apx:priors}

\subsection{Expert Priors}

We use Bayesian optimization (BO) \citep{mockus-tgo78a} with expert priors \citep{hvarfner2021pibo} as implemented in the NePS python package \citep{stoll2022neps}. Therefore, we set expert priors to guide the search. The priors in NePS are, in the continuous case, Gaussian distributions centered at a default value with the standard deviation determined via a confidential setting. We always use a  ``medium'' confidence and set the default value as described below.

\paragraph{Training Hyperparameters and Data Augmentation Strategy}
We set the defaults to the baseline for the training hyperparameters and data augmentation strategy. As no solarization is used in the SimSiam baseline and \citet{chen2021exploring} shows that adding solarize might improve the performance, we add solarize to the search space and set the user prior default to the values reported by \citet{chen2021exploring}.

\paragraph{RandAugment} For the RandAugment search space, we set the defaults according to the optimal data augmentation policy for CIFAR-10 following \citet{cubuk-cvpr20}.

\paragraph{SmartAugment} As we are the first applying SmartAugment to classification, we set the user prior defaults based on the optimal data augmentation policy for CIFAR-10 following \citet{cubuk-cvpr20}.

\paragraph{GroupAugment} For the GroupAugment config space, we set the default user priors to one augmentation per group. As only color and geometric augmentations occur in the baseline, we set the user prior defaults for these group probabilities to 0.5 and the other group probabilities user prior defaults to 0.

\subsection{Resources and Compute Budget}
For our Bayesian optimization runs, we allowed a budget of 50 evaluations. For CIFAR-10 and CIFAR-100, one configuration evaluation took $\approx$ 8h with one GeForce RTX 2080 Ti GPU, for DermaMNIST $\approx$ 10min with 1 GPU. We used 10 GPUs in parallel for CIFAR-10, 20 GPUs in parallel for CIFAR-100, and 5 GPUs in parallel for DermaMNIST. In order to take noise on the validation set during our HPO into account, we evaluate the best-performing configurations multiple times on the validation set.

\subsection{GroupAugment Comparative Study: RandAugment Search Space}
While SmartAugment and GroupAugment were designed with BO in mind, for RandAugment, which originally used Grid Search, we follow \citet{a15050165} and consider an extended search space for the number of operations between 1 and 15. \citet{a15050165} showed that a larger number of operations than 3 can be beneficial for the performance. See also Table~\ref{table:randaugment-config-spaces}.

\subsection{Dataset Splits} Since CIFAR-10 and CIFAR-100 \citep{krizhevsky-tech09a} do not provide a validation set, we split the training set and randomly sample a fixed validation set containing 10\% of the training data for our hyperparameter optimization. We use the entire training set for training for our final test evaluations of the best-performing validation configurations. For DermaMNIST \citep{medmnistv1,medmnistv2}, we have adopted the provided training, validation, and test split.

\newpage
\section{GroupAugment Details} \label{sec:appendix-groupaugment}

\begin{table}[ht!]
    \centering
    \caption{Details concerning the data augmentations from the groups. In our implementation, we use the data augmentations from the albumentations library \citep{buslaev2020albumentations}.}
    \begin{tabular}{ll}
    \toprule
    Group & Augmentation \\
    \midrule
    color & ColorJitter(brightness=0.4, contrast=0.4, saturation=0.4, hue=0.1) \\
    & ToGray() \\
    & Solarize() \\
    & Equalize() \\
    & ChannelShuffle() \\
    \midrule
    geometric & ShiftScaleRotate(interpolation=cv2.INTER\_CUBIC) \\
    & HorizontalFlip() \\
    \midrule
    non-rigid &  ElasticTransform(alpha=0.5, sigma=10, alpha\_affine=5, interpolation=cv2.INTER\_CUBIC)\\
    & GridDistortion(interpolation=cv2.INTER\_CUBIC) \\
    & OpticalDistortion(distort\_limit=0.5, shift\_limit=0.5, interpolation=cv2.INTER\_CUBIC) \\
    \midrule
    quality & GaussianBlur() \\
    & GaussNoise() \\
    \midrule
    exotic &  RandomGridShuffle()\\
    & Cutout(num\_holes=4) \\
    \bottomrule
    \end{tabular}
    \label{table:groupaugment_augmentations}
\end{table}

\newpage
\section{Individual Hyperparameter Analysis}\label{sec:hyperparameter-analysis}

Here, we individually analyze each hyperparameter and augmentation strategy parameter considered in Section~\ref{sec:study}. In particular, we plot density estimates for the values sampled in our search. We show density estimates for the best and worst-performing configurations (top and bad), all configurations (all), and configurations leading to collapsing. For categorical values, we show the sample distribution.

\subsection{Tuned SimSiam Data Augmentation Strategy}

\begin{figure}[H]
    \centering
    \includegraphics[width=\textwidth]{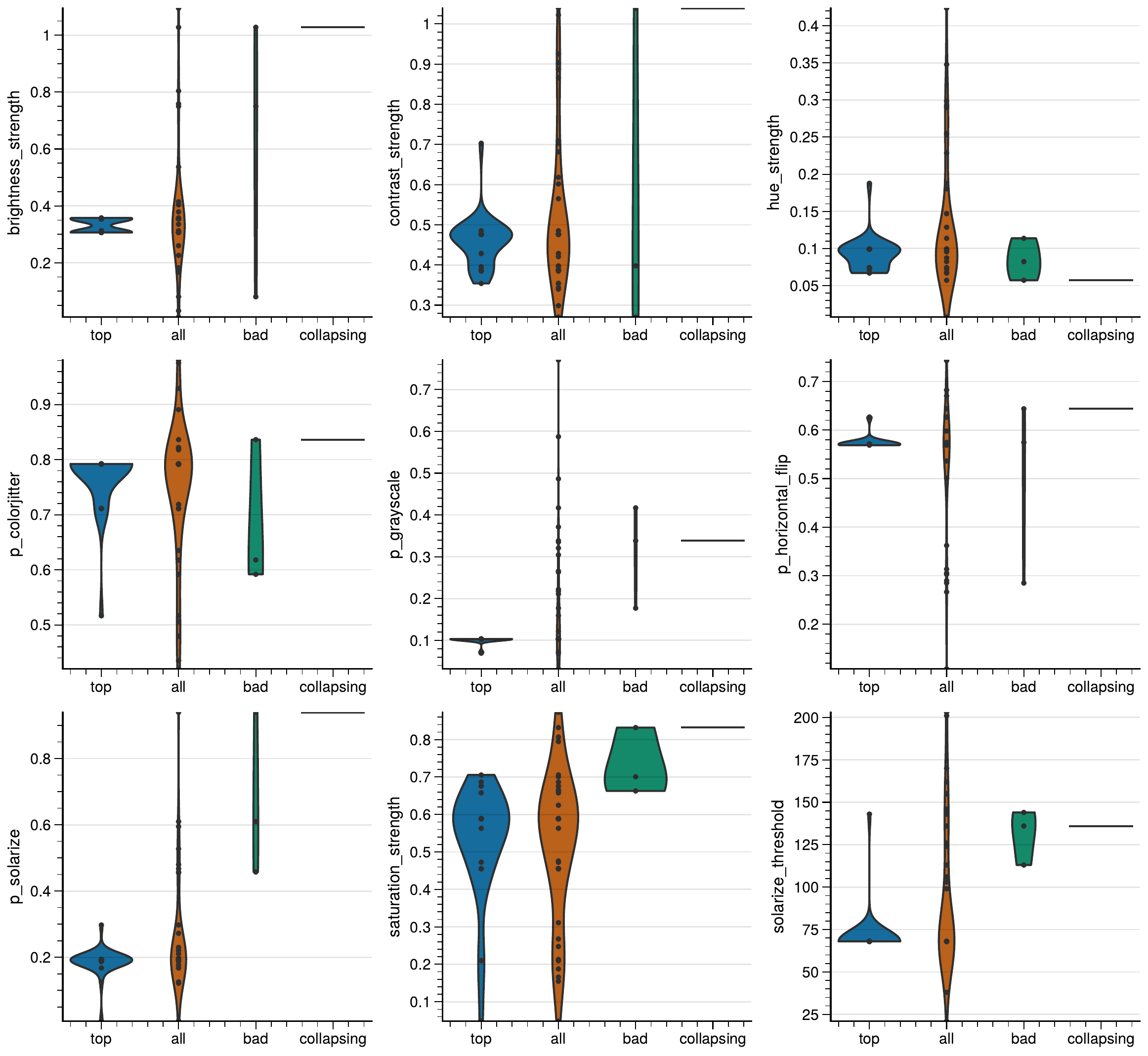}
    \caption{CIFAR-10 with SimSiam data augmentation search space.}
    \label{fig:cifar10-individual-simsiamdataaug}
\end{figure}

\begin{figure}[H]
    \centering
    \includegraphics[width=\textwidth]{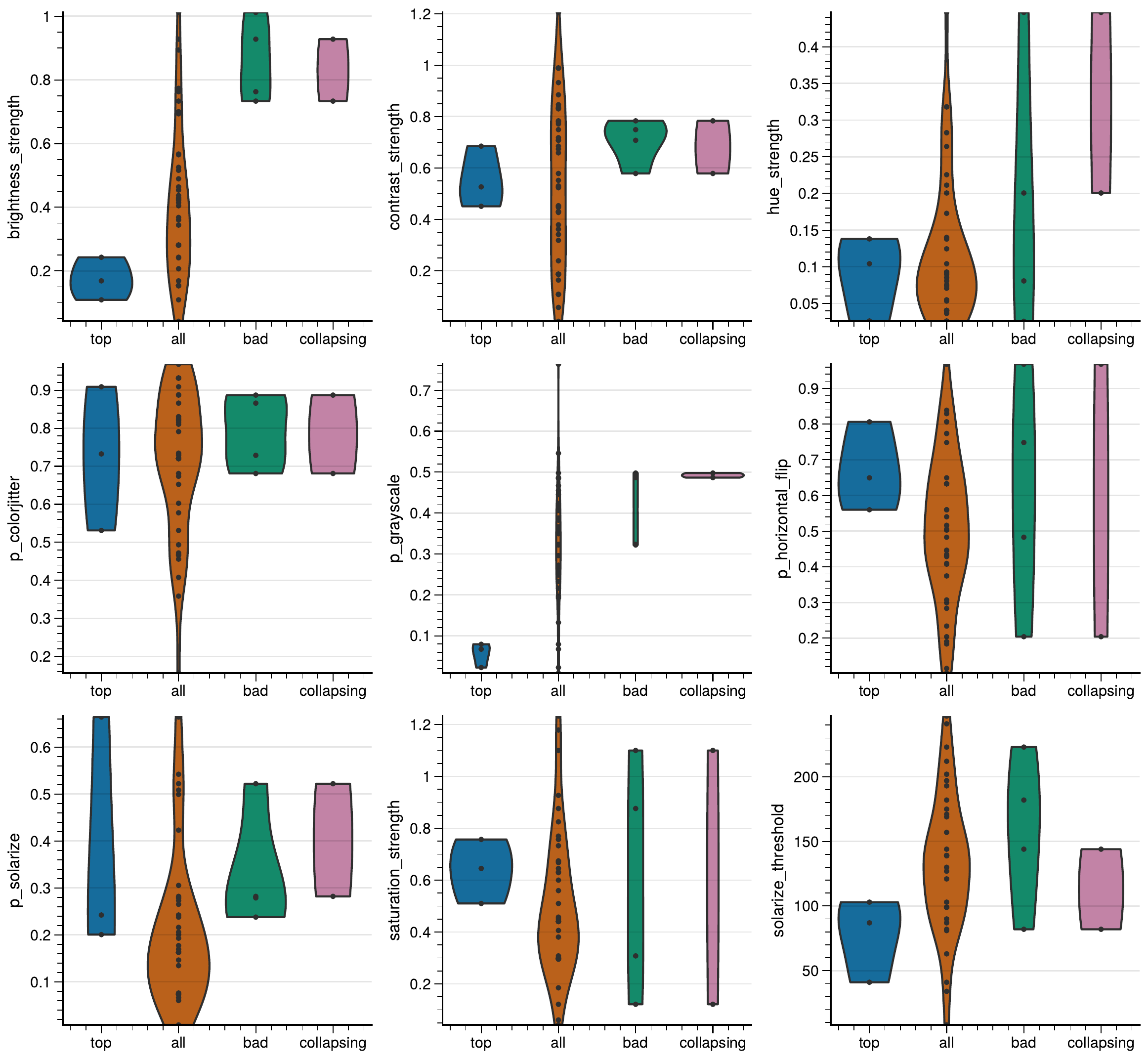}
    \caption{CIFAR-100 with SimSiam data augmentation search space.}
    \label{fig:cifar100-individual-dataaug}
\end{figure}

\begin{figure}[H]
    \centering
    \includegraphics[width=\textwidth]{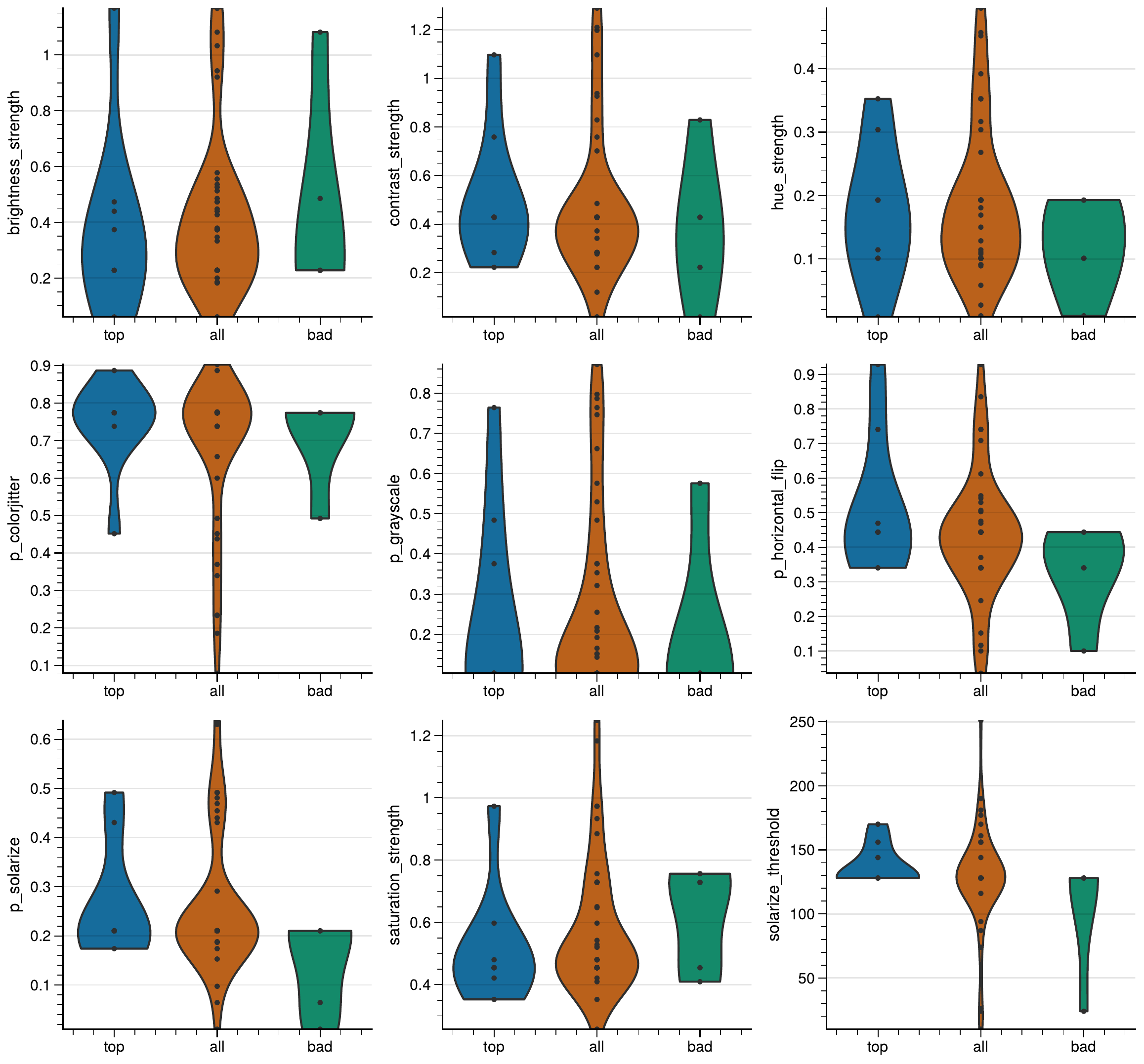}
    \caption{DermaMNIST with SimSiam data augmentation search space.}
    \label{fig:dermamnist-individual-dataaug}
\end{figure}

\newpage

\subsection{Tuned SimSiam Training Hyperparameters}

\begin{figure}[ht]
    \centering
    \includegraphics[width=\textwidth]{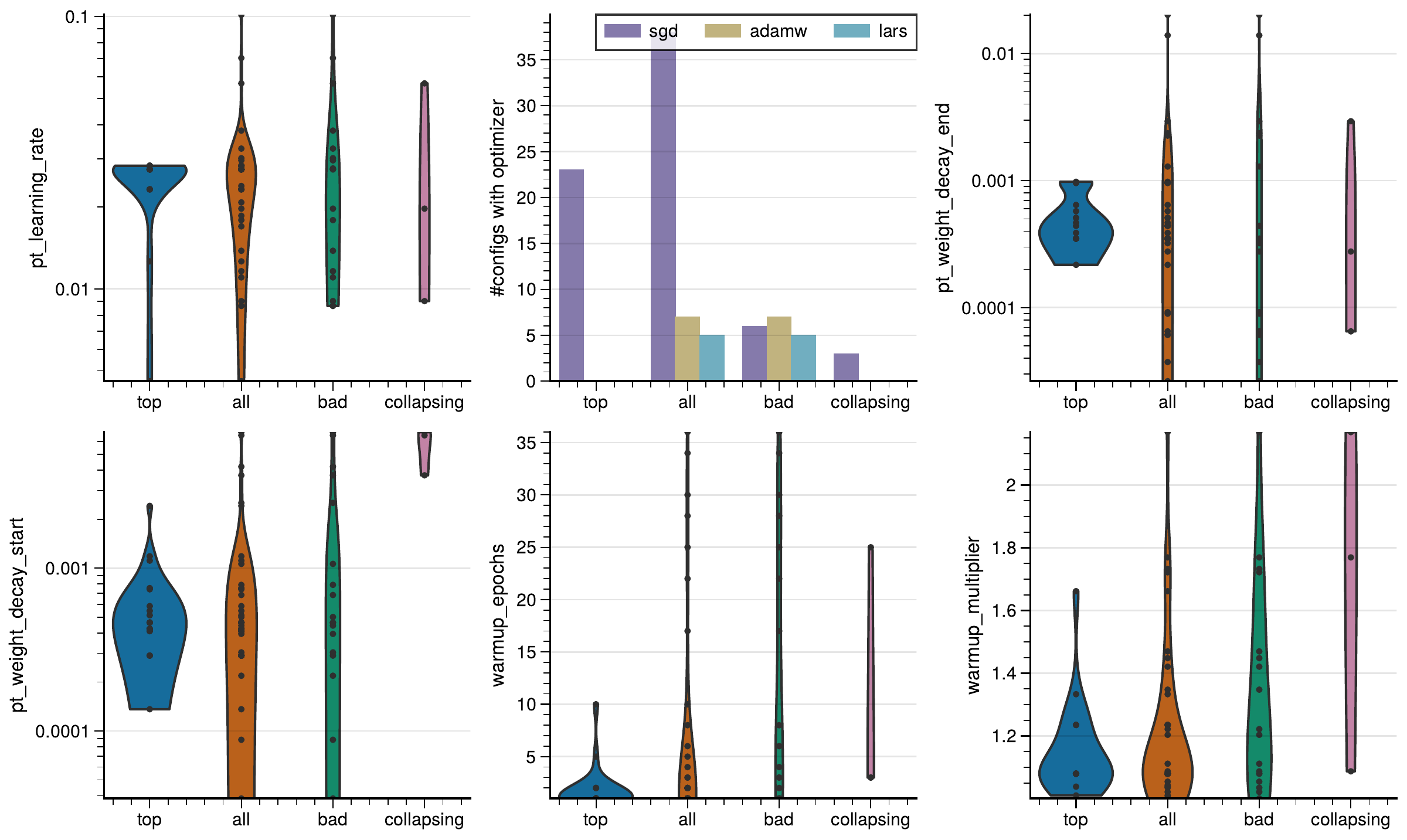}
    \caption{CIFAR-10 with SimSiam training hyperparameters search space.}
    \label{fig:training-hypers-violins-cifar10}
\end{figure}

\newpage

\begin{figure}[H]
    \centering
    \includegraphics[width=\textwidth]{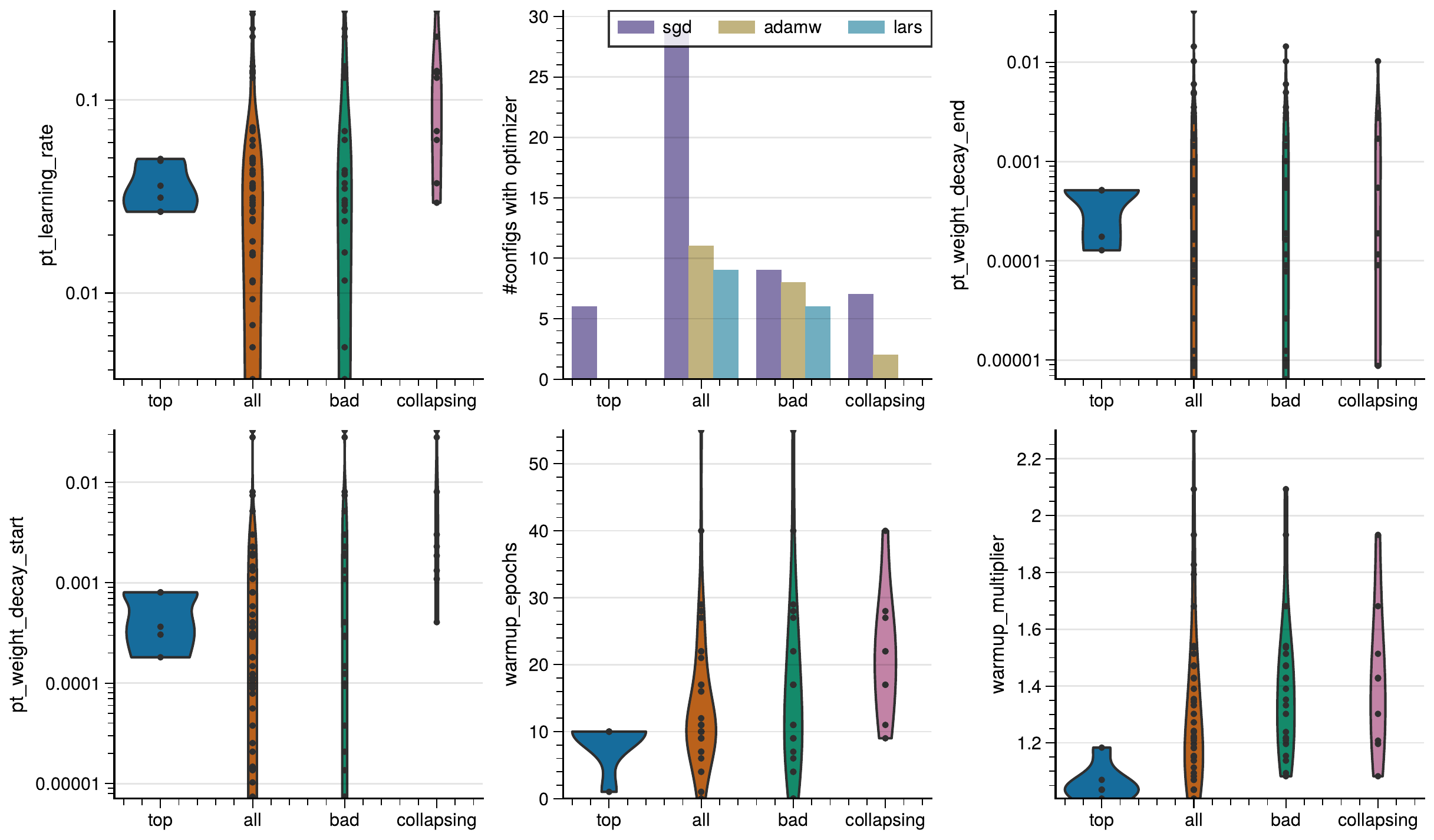}
    \caption{CIFAR-100 with SimSiam training hyperparameters search space.} 
    \label{fig:training-hypers-violins-cifar100}
\end{figure}

\begin{figure}[H]
    \centering
    \includegraphics[width=\textwidth]{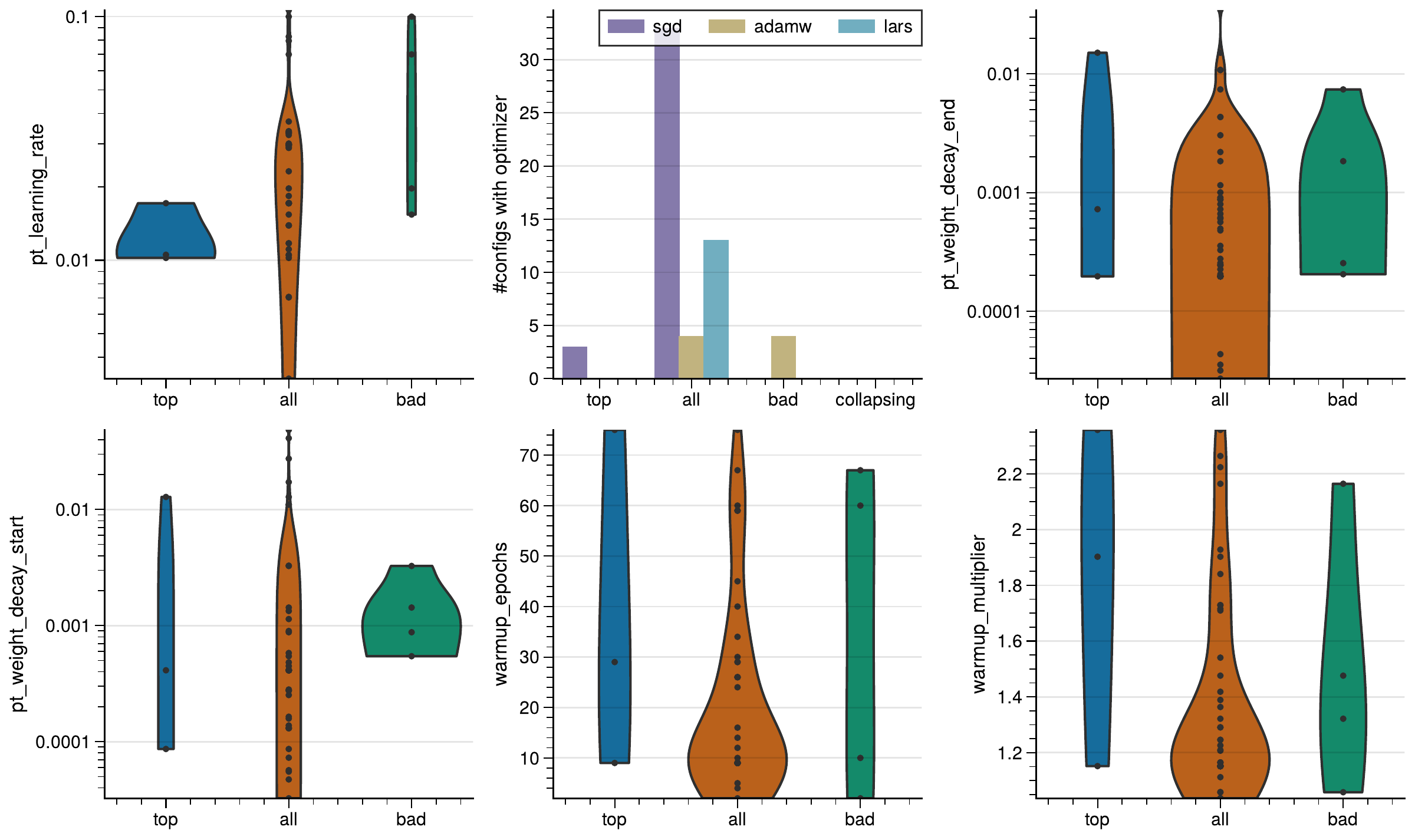}
    \caption{DermaMNIST with SimSiam training hyperparameters search space.}
    \label{fig:training-hypers-violins-dermamnist}
\end{figure}

\end{document}